\documentclass[runningheads]{llncs}

\usepackage{eccv}

\usepackage{eccvabbrv}

\usepackage{graphicx}
\usepackage{booktabs}
\usepackage{amsmath,bm}
\usepackage{multirow}
\usepackage{graphicx}
\usepackage[normalem]{ulem}
\useunder{\uline}{\ul}{}

\usepackage[accsupp]{axessibility}  

\usepackage[pagebackref,breaklinks,colorlinks,citecolor=eccvblue]{hyperref}
\usepackage{orcidlink}

\begin{document}

\title{LG-Traj: LLM Guided Pedestrian Trajectory Prediction}

\titlerunning{LG-Traj: LLM Guided Pedestrian Trajectory Prediction}

\author{Pranav Singh Chib\orcidlink{0000-0003-4930-3937} \and
Pravendra Singh\orcidlink{0000-0003-1001-2219 }}

\authorrunning{P.S. Chib et~al.}

\institute{ Indian Institute of Technology Roorkee, Uttarakhand, India \\
\email{\{pranavs\_chib, pravendra.singh\}@cs.iitr.ac.in}
}

\maketitle

\begin{abstract}

Accurate pedestrian trajectory prediction is crucial for various applications, and it requires a deep understanding of pedestrian motion patterns in dynamic environments. However, existing pedestrian trajectory prediction methods still need more exploration to fully leverage these motion patterns. This paper investigates the possibilities of using Large Language Models (LLMs) to improve pedestrian trajectory prediction tasks by inducing motion cues. We introduce LG-Traj, a novel approach incorporating LLMs to generate motion cues present in pedestrian past/observed trajectories. Our approach also incorporates motion cues present in pedestrian future trajectories by clustering future trajectories of training data using a mixture of Gaussians. These motion cues, along with pedestrian coordinates, facilitate a better understanding of the underlying representation. Furthermore, we utilize singular value decomposition to augment the observed trajectories, incorporating them into the model learning process to further enhance representation learning. Our method employs a transformer-based architecture comprising a motion encoder to model motion patterns and a social decoder to capture social interactions among pedestrians. We demonstrate the effectiveness of our approach on popular pedestrian trajectory prediction benchmarks, namely ETH-UCY and SDD, and present various ablation experiments to validate our approach.

\end{abstract}

\section{Introduction}
Trajectory prediction is the process of anticipating pedestrian's future motions based on their past motion. This task is crucial for self-driving cars, behavioural analysis, robot planning, and other autonomous systems. When forecasting a pedestrian's future trajectory, a wide range of trajectories can be possible, and learning such varied spatio-temporal representations of trajectories is a major challenge in pedestrian trajectory prediction. The trajectories of individual pedestrians are influenced by various factors, including their inherent motion characteristics and the social interactions of neighbouring pedestrians. Previous works have employed recurrent networks \cite{alahi2016social,robicquet2016learning} to model the underlying motion pattern of each pedestrian. Additionally, graph-based methods \cite{mohamed2020social,salzmann2020trajectron++,bae2023set,shi2021sgcn,huang2019stgat,mendieta2021carpe} predict the motion pattern of agents using graph structures, where vertices represent pedestrians and edges represent their interactions. Generative-based approaches, such as GANs \cite{sophie19, hu2020collaborative,gupta2018social} and VAEs \cite{Xu_2022_CVPR,lee2022muse,xu2022dynamic,mangalam2020not}, model the distribution of plausible future motion. The abovementioned methods leverage the spatio-temporal information from the given data to understand pedestrian motion dynamics. Building on these efforts, we investigate the possibility of improving pedestrian trajectory prediction task by leveraging motion cues generated by a large language model (LLM).

\begin{figure*}[!t]
    \centering
    \includegraphics[scale=.113]{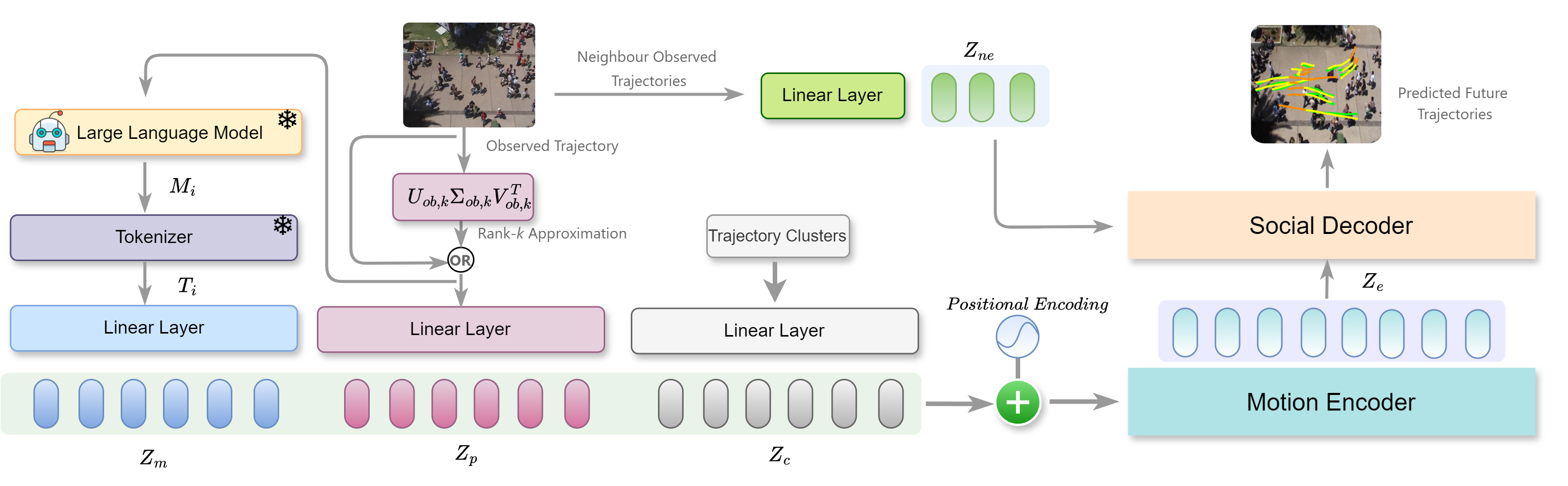}

\caption{The overview of our proposed LG-Traj involves taking multiple inputs including past motion cues, past observed trajectory, and future motion cues. First, we augment the given observed trajectory using rank-\textit{k} approximation via singular value decomposition (SVD). Then for the subsequent steps, we either use the original past observed trajectory or augmented past observed trajectory. Next, we generate past motion cues ($M_i$) from LLM using the past observed trajectory ($X_i$) of the $i^{th}$ pedestrian. Tokenizer output ($T_i$) is generated from $M_i$ by the tokenizer. Past motion cues embedding ($Z_m$) is obtained by a linear transformation of $T_i$. Past trajectory embedding ($Z_p$) is obtained by a linear transformation of $X_i$. Cluster embedding $Z_c$ is obtained by a linear transformation of trajectory clusters. Trajectory clusters are generated by clustering future trajectories of training data using a mixture of Gaussians. Positional encoding is added to the concatenated embeddings ($Z_m, Z_p, Z_c$), and the result is passed as an input to the motion encoder to model the motion patterns. The embedding generated by the motion encoder ($Z_e$) along with neighbour embedding ($Z_{ne}$) is passed as an input to the social decoder to predict future trajectories.}

    \label{fig_main_lhtraj}
\end{figure*}

In this work, we present a novel approach called \textbf{L}LM \textbf{G}uided \textbf{Traj}ectory prediction (LG-Traj). Our approach effectively incorporates motion cues, along with the past observed trajectories, to predict future trajectories (see Fig.~\ref{fig_main_lhtraj}). We use a motion encoder to integrate spatio-temporal motion patterns and a social decoder to capture social interactions among pedestrians for accurate trajectory prediction. We utilize LLM to generate past motion cues present in pedestrian past trajectories. Additionally, we utilize future motion cues present in pedestrian future trajectories by clustering future trajectories of training data using a mixture of Gaussians. Specifically, past motion cues, past observed trajectory, and future motion cues are utilized by the motion encoder of the transformer to model the motion patterns (see Fig.~\ref{fig_main_lhtraj}). Furthermore, the social decoder of the transformer uses the social interactions of neighbouring pedestrians along with the embedding generated by the motion encoder to generate socially plausible future trajectories. Additionally, to effectively model the past trajectories, we augment the observed trajectories by singular value decomposition and incorporate them into the training process to further enhance representation learning. We also explore several existing data augmentation techniques in our approach, but we obtain suboptimal results. More details are provided in the supplementary material. Our extensive experimentation on popular pedestrian benchmark datasets, namely ETH-UCY \cite{lerner2007crowds,pellegrini2009you} and SDD \cite{robicquet2016learning}, demonstrates the effectiveness of our proposed approach. We also present various ablation experiments to validate our approach.

\section{Related Work}

\subsection{Trajectory Prediction}

Trajectory prediction involves forecasting the trajectory at future timestamps given past observations. Since future states can evolve from the current state, sequence-to-sequence modelling approaches can be used to model these trajectory sequences. Recurrent Neural Networks (RNNs) \cite{alahi2016social, huang2019stgat} and Long Short-Term Memory networks (LSTM) \cite{hochreiter1997long} have made significant progress in sequence prediction tasks. These architectures have been utilized to learn the temporal patterns of pedestrian trajectories. Moreover, LSTM networks \cite{ivanovic2019trajectron, vemula2018social} construct spatio-temporal networks capable of representing structured sequence data. However, it is worth noting that RNN-based models may encounter issues like gradient vanishing or explosion under specific circumstances. Since the movement of pedestrians is uncertain, there may exist variations in future trajectories. To capture this variation in future trajectories, deep generative models such as Generative Adversarial Network (GAN) \cite{sophie19, hu2020collaborative,gupta2018social}, Variational Auto-Encoder (VAE) \cite{Xu_2022_CVPR,lee2022muse,xu2022dynamic,mangalam2020not}, normalizing flow \cite{bhattacharyya2019conditional}, and diffusion-based models \cite{mao2023leapfrog,gu2022stochastic} are used. Transformers demonstrated satisfactory performance in trajectory prediction \cite{girgis2022latent,yuan2021agentformer,zhou2023query,yu2020spatio} and have been frequently used to model long-range relationships. Some work \cite{lv2023ssagcn,sekhon2021scan} uses graph neural network-based techniques by building a graph structure containing pedestrian nodes and interaction edges for trajectory prediction. The graph-structured pedestrian characteristics are updated using transformers \cite{yuan2021agent,gu2022mid,monti2022stt,wen2022socialode,wong2022v2net}, graph convolutional networks \cite{lv2023ssagcn,bae2023set}, and graph attention networks \cite{huang2019stgat,kosaraju2019social,shi2021sgcn,bae2022npsn}. Despite significant progress in trajectory prediction, as mentioned above, there is a need for further exploration to leverage motion cues effectively. Our approach leverages motion cues from the LLM to move forward in this direction.

\subsection{Large Language Model}

Large language models \cite{lewis2019bart,achiam2023gpt,touvron2023llama} have started to be used in scene understanding tasks, including object localization \cite{dewangan2023talk2bev}, scene captioning \cite{aydemir2023adapt}, and visual question answering \cite{sima2023drivelm,marcu2023lingoqa}. For example, in autonomous driving, DriveLikeHuman \cite{fu2023drive} uses LLMs to create a new paradigm that mimics how humans learn to drive. Similarly, GPT-Driver \cite{mao2023gpt} uses GPT-3.5 to improve autonomous driving with reliable motion planning. Parallel to this, SurrealDriver \cite{jin2023surrealdriver} builds an LLM-based driver agent with memory modules that mimic human driving behaviour to comprehend driving scenarios, make decisions, and carry out safe actions. ADAPT \cite{aydemir2023adapt} offers explanations in driving captions to understand every stage of the decision-making process involved in autonomous vehicle control. LLM planning capabilities are also being used in robot navigation \cite{huang2023visual,parisi2022unsurprising}, where natural language commands are translated into navigation goals. In pedestrian trajectory prediction, there has not been much effort to utilize the capabilities of large language models. In contrast, we leverage the capabilities of LLM by training the trajectory prediction model with motion cues generated by LLM to understand the underlying trajectory motion patterns better.

\section{Method}

\subsection{Problem Definition}

Formally, the observation trajectory with length $T_{\text{ob}}$ can be represented as $\bm{X}_i = \{ (x_i^t, y_i^t) \, |~t \in [1, \dots, T_{\text{ob}}] \}$, where $(x_i^t, y_i^t)$ is the spatial coordinate of a pedestrian $i$ at $t^{th}$ time. Furthermore, ground truth future trajectory for prediction time length $T_{\text{pred}}$ can be defined as $\bm{Y}_i = \{ (x_i^t, y_i^t) \, | \, t \in [T_{\text{ob}} + 1, \ldots, T_{\text{pred}}] \}$. 
The number of pedestrians in the scene is represented by $N$, and $i \in N$ denotes the pedestrian index. The goal of trajectory prediction is to generate future trajectories $(\bm{\hat{Y}_i})$ that closely approximate the ground truth trajectory $(\bm{{Y}_i})$.

\subsection{Trajectories Augmentation} \label{sec:approxima}
We first construct the trajectory matrix X by stacking all pedestrian observations from a training batch. We then utilize Singular Value Decomposition (Eq.~\ref{eq:svd}) followed by the rank-\textit{k} approximation (Eq.~\ref{eq:best_rank_k}) to obtain the augmented trajectories.

\begin{equation}
   \bm{\text{X}} = \bm{U}_{\!ob} \bm{\mathit{\bm{S}}}_{\!ob} \bm{V}^\top_{\!ob}
    \label{eq:svd}
\end{equation}

where $\bm{U}_{ob}\!\!=\![\bm{u}_1, \cdots, \bm{u}_L]$ and $\bm{V}_{ob}\!\!=\![\bm{v}_1, \cdots, \bm{v}_N]$ are orthogonal matrices and $\bm{\mathit{\bm{S}_{ob}}}$ is a diagonal matrix, consisting of singular values $\sigma_1\!\geq\!\sigma_2\!\geq\!\cdots\!\geq\!\sigma_r\!>\!0$. Here, $L=2\!\,\times\!\,T_{ob}$ for observed pedestrian trajectory. $N$ is the number of pedestrians, and $r$ is the rank of X. We use the rank-\textit{k} approximation of X by using the first $k$ singular vectors as shown below:
\begin{equation}
    \tilde{\bm{\text{X}}} = \bm{U}_{\!ob,k} \,\bm{S}_{\!ob,k} \,{\bm{V}^\top_{\!ob,k}}
    \label{eq:best_rank_k}
\end{equation}

Where $\bm{U}_{\!ob,k} = [\bm{u}_1, \cdots, \bm{u}_k$], $\bm{S}_{\!ob,k} = \text{diag}(\sigma_1, \ldots, \sigma_k)$, $\bm{V}_{\!ob,k} = [\bm{v}_1, \ldots, \bm{v}_k]$. Also, $\tilde{\bm{\text{X}}}$ is the approximated matrices obtained from the best rank-$k$ approximation of $\bm{\text{X}}$, containing only the $k$ most significant singular values. This approximation allows us to preserve critical information with minimal loss of information, as shown below:

\begin{equation} \label{eq_info_loss}
    \text{Information Loss} = 1 - \frac{\sum_{t=1}^{k} \sigma_t}{\sum_{t=1}^{r} \sigma_t}
\end{equation}
  
Here, $r$ is the total number of singular values, and $\sigma_t$ represents the $t^{th}$ singular value. We augment the given past observed trajectory using rank-\textit{k} approximation via singular value decomposition (SVD). Throughout the remainder of the paper, we exclusively present formulations using the original past observed trajectories for the sake of simplicity. However, it is important to note that the same formulation is applicable to augmented past observed trajectories.

\begin{figure*}[!t]
    \centering
    \includegraphics[scale=.12]{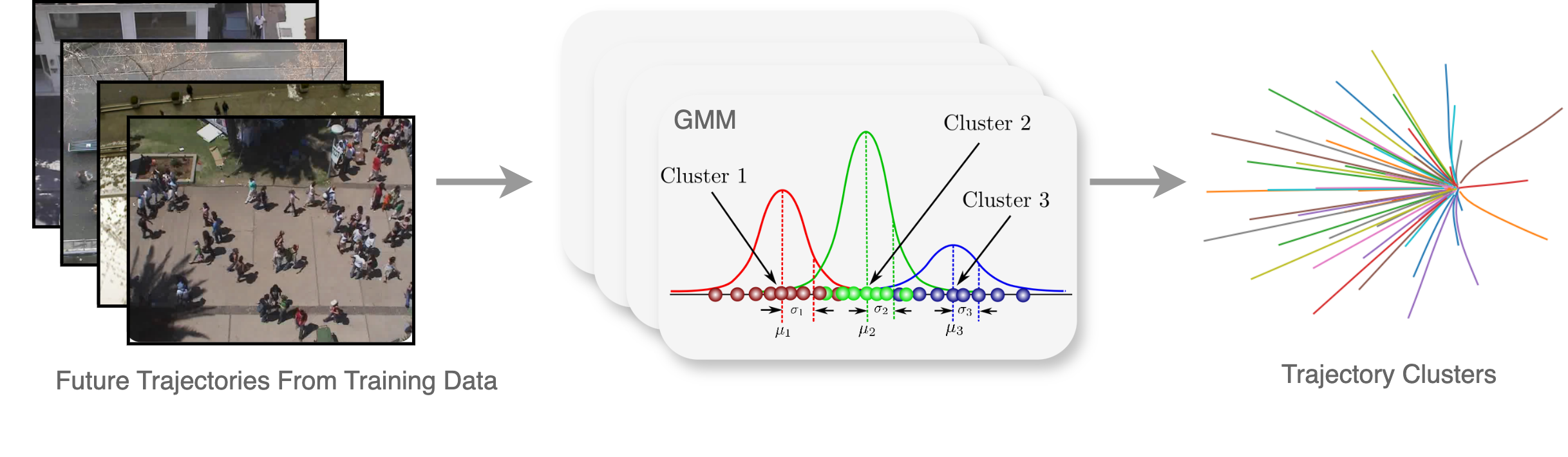}
\caption{Illustration of the mixture of Gaussians, where each Gaussian represents a diverse cluster of trajectories.}
    \label{fig_gmm_cluster}
\end{figure*}

\subsection{Clustering using Mixture of Gaussians} \label{sec:mixedgaussin}
We utilize the Gaussian Mixture Model to model the diverse future motion cues from the training data (see Fig.~\ref{fig_gmm_cluster}). Specifically, we construct mixed Gaussian clusters, which are a combination of $\mathcal{C}$ Gaussians ($\{ \mathcal{N}(\mu_c, \sigma^2_c) \}, (1 \leq c \leq \mathcal{C})$). These multiple Gaussians represent different motion cues present in future trajectories (i.e., linear motion, curved motion, etc.). To generate clusters, we first preprocess the future trajectories by translating all starting coordinates to the origin, followed by rotation to the positive zero-degree direction. Then, we cluster similar motion behaviours of future trajectories into $\mathcal{C}$ clusters with their respective clusters means $\mu = \{ \mu_{c_j} \}_{j=1,\ldots,\mathcal{|C|}}$. For instance, if there are 50 trajectory clusters, then \( \mathcal{C} \in \mathbb{R}^{50 \times T_{\text{pred}} \times 2} \) and clusters $\{ c_1, c_2, \dots c_{50}\}$. 
The mixture of Gaussians are shown below:

\begin{equation}
    \mathbf{M} = \sum_{j=1}^{\mathcal{|C|}} \mathbf{W}_{c_j} ~\mathcal{N}(\boldsymbol{\mu}_{c_j}, \boldsymbol{\sigma}^2_{c_j})
\end{equation}

Here, $\mathbf{W}_{c_j}$ is the learnable parameter for assigning weights to the clusters, while $\mu_{c_j}$ and $\sigma^2_{c_j}$ represent the mean and variances of the Gaussians. The output clusters signify diverse motion cues in pedestrian future trajectories. 

\subsubsection{Soft Trajectory Probability}  \label{sec:soft_prob}

LG-Traj outputs the predicted trajectories along with their corresponding probabilities, which is beneficial for estimating the uncertainty associated with each prediction. For instance, lower probabilities indicate that the model is less confident about the prediction. Specifically, following the nearest neighbour hypothesis \cite{1053964}, the trajectory cluster $c_{j}$ closer to the ground truth is the most likely one, i.e., owning the maximum probability. The soft probability ${p}_i$ for $i^{th}$ pedestrian is computed by taking the negative squared Euclidean distance between the ground truth trajectory and the cluster centre, normalizing it, and then applying the softmax as shown below:

\begin{equation}
    p_i = \frac{e^{-\| {Y_i} - c_{j} \|_2^2}}{\sum_{j=1}^{|\mathcal{C}|} e^{-\| {Y_i} - c_j \|_2^2}}
\end{equation}

\begin{figure*}[!t]
    \centering
    \includegraphics[scale=.09]{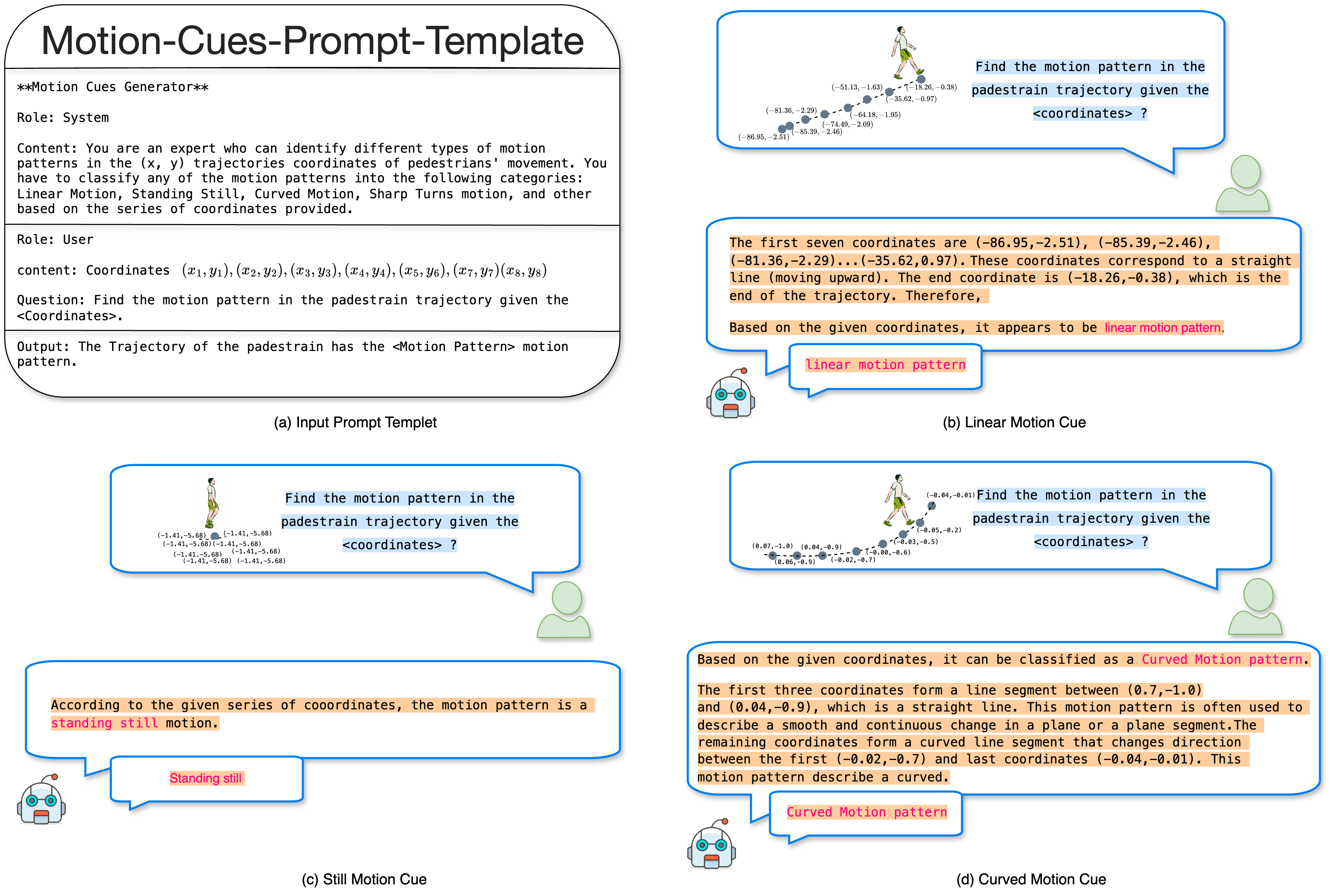}
   
\caption{Illustration of input prompt and examples of motion cues generation from the LLM. We present three different examples where the LLM correctly identifies the underlying trajectory motion pattern, such as linear motion, curved motion, and standing still, based on the coordinates provided as input to the LLM.}
    \label{fig_prompttemplet}
\end{figure*}

\vspace{-10pt}
\subsection{Past Motion Cues generated by LLM} \label{sec:llmmotionclue}

\subsubsection{Prompt Engineering}
Directly feeding the trajectory coordinates to the LLMs does not always produce the expected output, such as the motion cues we desire. In this prompt engineering step, we guide the LLMs to generate the motion cues given the observed past trajectory of the pedestrian. Our prompt design is shown in Figure~\ref{fig_prompttemplet}(a). \newline

We adapted the chat template for our task to generate the motion cues. Our simple prompt can infer the motion cues when the trajectory coordinates sequence is fed to the LLM. We begin by specifying the system's role, where we provide a precise description of the system and its task, which is to identify motion patterns in pedestrian movements. Next, we specify the format of the user role and input data to the LLM. This simple template formatting ensures that the LLM correctly aligns the coordinates with the motion cues. The generated cues are shown in Figure~\ref{fig_prompttemplet}. The first example contains the linear motion trajectory of the pedestrian, and the LLM correctly identifies the pattern, followed by the stationary and curved motion of the pedestrian.

\subsubsection{Generation}

Given a set of observed coordinates for the $i^{th}$ pedestrian trajectory $X_i$. A Language Model ($f_{\text{LLM}}$) generates past motion cues \(M_i\) containing the motion patterns of the pedestrian:
\begin{equation}
    M_i = f_{\text{LLM}}(\bm{X}_{i})
\end{equation}

The past motion cues \(M_i\) for the $i^{th}$ agent is transformed into an tokenized output embedding vector \({T}_{i}\) using a tokenizer:
\begin{equation}
    T_{i} = g_{\text{ST}}(M_i)
\end{equation}

Here, \( M_i \) is the past motion cues generated by the LLM for the $i^{th}$ pedestrian and $g_{\text{ST}}$ is the tokenizer.

\subsection{Motion Embedding} \label{sec:motionembed}

Our approach utilizes linear layers to embed the past motion cues (Eq. \ref{eq_past_motion_clues}), future motion cues (Eq. \ref{eq_cluster}), and past observed trajectory (Eq. \ref{eq_pastcoo}). Additionally, we incorporate a positional encoding to model the temporal representation from the data to understand pedestrian motion dynamics. Positional encoding enables the model to capture long-term temporal dependencies within the underlying data. The embeddings from the LLM are tokenized and then transformed to yield the past motion cues embedding, as given below:

\begin{equation}\label{eq_past_motion_clues}
Z_m = \mathcal{F}_m (\mathbf{T}, \mathbf{W}_m)
\end{equation} 

Here, $\mathcal{F}_m(\cdot, \cdot)$ denotes a linear layer with trainable parameter matrix $\mathbf{W}_m$. $\text{T} \in \mathbb{R}^{B \times E_{s}}$, where $B$ is the batch size and $E_{s}$ is the output dimension of the tokenizer. The past motion cues embedding is $Z_m \in \mathbb{R}^{\text{B} \times M_d}$, where $M_d$ is the output dimension. Next, the cluster embedding is given as:

\begin{equation}\label{eq_cluster}
Z_c = \mathcal{F}_c ({\mathbf{\mathcal{C}}}, \mathbf{W}_c),
\end{equation}

Here, $Z_c \in \mathbb{R}^{|\mathcal{C| }\times O_c}$ represents cluster embedding. $O_c$ is the output dimension of the linear layer and $|C|$ is the number of clusters. Finally, the past trajectory embedding is given as:
\begin{equation}\label{eq_pastcoo}
Z_p = \mathcal{F}_p ({\mathbf{X}}, \mathbf{W}_p),
\end{equation}
Here, the embedding $Z_p$ denotes the past trajectories embedding, and trajectory matrix $\mathbf{X} \in \mathbb{R}^{B \times T_{ob} \times 2}$. After obtaining all the embeddings, the past motion cues embedding, cluster embedding, and past trajectory embedding are concatenated and added with positional encoding to get the motion embedding $Z_{f} = \text{\textit{concat}}(Z_m, Z_c, Z_p) + PE$, which is then passed as an input to the motion encoder. Here, $Z_{f} \in \mathbb{R}^{B \times |\mathcal{C}| \times {M}_e} $ and $PE$ is a tensor containing positional encoding information.

\vspace{-5pt}
\subsection{Motion Encoder} \label{sec:motionencoder}

The motion encoder is designed to model the spatio-temporal motion patterns in pedestrian trajectories. Specifically, the encoder consists of multiple identical layers (i.e., $L$ number of layers), each layer consisting of a multi-head self-attention and feed-forward network. The multi-head self-attention is given as:

\begin{align}
    \text{MultiHead}(Q, K, V) &= \text{Concat}(\text{head}_1, \ldots, \text{head}_h)W^O \\
    \text{head}_i &= \text{Attention}(QW_i^Q, KW_i^K, VW_i^V) \\
    \text{Attention}(Q, K, V) &= \text{Softmax}\left(\frac{QK^T}{\sqrt{d_k}}\right)V
\end{align}

where $d_k$ is the dimension of $K$ or $V$. $Q, K, V$ represent the query, key, and value matrices, respectively, and $W_i^Q, W_i^K, W_i^V$ are learnable linear transformations for the $i^{th}$ head. $W^O$ is another learnable linear transformation for the projection of output. The feed-forward network is given as:

\begin{align}
    \text{FFN}(x) &= \text{ReLU}(xW_1 + b_1)W_2 + b_2
\end{align}

where $W_1, W_2$ are learnable weight matrices, and $b_1, b_2$ are bias vectors for the feed-forward network. The final output of the encoder $Z_e$ is obtained by passing the motion embedding $Z_f$ through $L$ encoder layers:

\begin{align}
    Z_e &= \text{EncoderLayer}(Z_f)
\end{align}

where $Z_e \in \mathbb{R}^{B\times 1\times embed\_size}$, $embed\_size$ is the size of the embedding, $B$ is the batch size. 

\vspace{-5pt}
\subsection{Social Decoder} \label{sec:social_decoder}

The social decoder combines the pedestrian's motion patterns with the social interactions of neighbour pedestrians. Neighbour embedding ($Z_{ne}$) is obtained by applying linear transformation of neighbour past observed trajectories. Neighbour embedding ($Z_{ne}$) along with the output $Z_e$ from the motion encoder are fed into the decoder to forecast future trajectories. The query embedding represents the current pedestrian ($\mathbf{Q} \in \mathbb{R}^{B \times 1 \times\text{embed\_size}}$). Key and value embeddings represent neighbouring pedestrians ($\mathbf{K}, \mathbf{V} \in \mathbb{R}^{B \times N \times\text{embed\_size}}$).

Through self-attention, the decoder weighs various neighbour interactions in relation to the current pedestrian. Furthermore, the resulting embeddings from the decoder represent the predicted future trajectory of the pedestrian. Along with the predicted future trajectory, we also predict the probability that estimates the uncertainty associated with each prediction.

\begin{align}
    Z_n, ~\hat{p}_n &= \text{DecoderLayer}(Z_e, Z_{ne})
\end{align}

Here, $Z_e$ is the output of the encoder and $Z_{ne}$ is the neighbour embedding. $Z_n \in \mathbb{R}^{B \times num \times T_{pred} \times 2}$ is the output of the decoder, where $num$ is the number of trajectories to be predictions at inference time. $\hat{p}_n$ is the probabilities associated with predicted trajectories. 

During test time, we generate past motion cues from LLM using the observed trajectory of test data. For future motion cues, we utilize the same trajectory clusters that were used during training. Finally, our model leverages past motion cues, the observed trajectory from test data, and future motion cues to predict the future trajectory for the test data.

\subsection{Training Loss}
Our training loss consists of two components: trajectory prediction loss ($\mathcal{L}_{\text{traj}}$), and loss for the corresponding trajectory probabilities ($\mathcal{L}_{\text{prob}}$). For trajectory prediction, we use the Huber loss between the predicted trajectory ($\hat{Y}_i$) and the ground truth trajectory (${Y}_i$). For probabilities, we use Cross-Entropy loss between the ground truth probability and the predicted probability. The trajectory prediction loss is defined as:

\begin{equation}
\mathcal{L}_{\text{traj}} = \frac{1}{N} \sum_{i=1}^{N} \text{Huber}(Y_i, \hat{Y}_i)
\end{equation}
where $\text{Huber}(Y_i, \hat{Y}_i)$ with $\delta$ threshold is defined as:

\begin{equation}
\text{Huber}(Y_i, \hat{Y}_i) = \begin{cases} 
\frac{1}{2} (Y_i - \hat{Y}_i)^2 & \text{if } |Y_i - \hat{Y}_i| \leq \delta \\
\delta (|Y_i - \hat{Y}_i| - \frac{1}{2} \delta) & \text{otherwise}
\end{cases}
\end{equation}

The loss for the trajectory probabilities is cross-entropy loss $L_{\text{prob}}$ which is defined as:

\begin{equation}
\mathcal{L}_{\text{prob}} = -\frac{1}{N} \sum_{i=1}^{N}  p_{i} \log(\hat{p}_{i})
\end{equation}

where $p_{i}$ is the ground truth trajectory probability (ref Section \ref{sec:soft_prob}) and $\hat{p}_{i}$ is the predicted probability. The overall loss function $\mathcal{L}_{total}$ is defined below.
\begin{equation}
    \mathcal{L}_{total} =  \mathcal{L}_{\text{traj}} + \mathcal{L}_{\text{prob}}
\end{equation}

\begin{table}[!t]
\centering
\caption{Details about the SDD, ETH, HOTEL, UNIV, ZARA1, and ZARA2 datasets. The \emph{Scenes} column indicates the number of scenarios during which trajectories were recorded.}
\label{tab_dataset}
\resizebox{\textwidth}{!}{%
\begin{tabular}{c|c|c|c}
\hline
Dataset & Scenes & Number of agents & Description \\ \hline
SDD  & 20 scenes                 & 11,000 Agents               & Past frames: 8 (3.2s), and predicted frames: 12 (4.8s)                 \\ \hline
ETH  & \multirow{5}{*}{4 scenes} & \multirow{2}{*}{750 Agents} & \multirow{2}{*}{Past frames:8 (3.2s), and predicted frames: 12 (4.8s)} \\ \cline{1-1}
HOTEL   &        &                  &             \\ \cline{1-1} \cline{3-4} 
UNIV &                           & \multirow{3}{*}{786 Agents} & \multirow{3}{*}{Past frames:8 (3.2s), and predicted frames: 12 (4.8s)} \\ \cline{1-1}
ZARA1   &        &                  &             \\ \cline{1-1}
ZARA2   &        &                  &             \\ \hline
\end{tabular}%
}
\end{table}

\section{Experiments}

\subsection{Experimental settings}
\subsubsection{Datasets}
To validate our proposed approach, we conduct experiments on two benchmark datasets (Table \ref{tab_dataset}): the Stanford Drone Dataset (SDD) \cite{robicquet2016learning} and ETH-UCY dataset \cite{lerner2007crowds,pellegrini2009you}. ETH-UCY is a widely used benchmark dataset for predicting pedestrian trajectories, consisting of the trajectories of 1,536 pedestrians in four distinct scenarios split into five subsets: ETH, HOTEL, UNIV, ZARA1, and ZARA2. These scenarios include various scenes such as roads, intersections, and open areas. SDD is also a benchmark dataset providing bird's-eye-view perspectives of pedestrian trajectory prediction, with 5,232 trajectories from eight distinct scenarios. We used the same experimental settings for both ETH-UCY and SDD, with an observed trajectory length of 3.2 seconds (8 frames) and a predicted trajectory length of 4.8 seconds (12 frames) as used by the compared methods for fair comparison.

\subsubsection{Evaluation Metrics}
To assess the effectiveness of our method, we use widely used evaluation metrics for trajectory prediction, such as Average Displacement Error (ADE) and Final Displacement Error (FDE). ADE measures the average difference ($l_2$ distance) between the predicted and actual future positions of a pedestrian for all prediction time steps.
\begin{equation}
ADE = \frac{1}{T_{\text{pred}} - T_{\text{ob}}} \sum_{t=T_{\text{ob}} + 1}^{T_{\text{pred}}} \sqrt{(x_i^t - \hat{x}_i^t)^2 + (y_i^t - \hat{y}_i^t)^2}
\end{equation}

FDE measures the difference ($l_2$ distance) between the predicted future endpoint position and the actual future endpoint position.
\begin{equation}
FDE = \sqrt{(x_i^{T_{\text{pred}}} - \hat{x}_i^{T_{\text{pred}}})^2 + (y_i^{T_{\text{pred}}} - \hat{y}_i^{T_{\text{pred}}})^2}
\end{equation}

Here, $T_{\text{ob}}$ \& $T_{\text{pred}}$ are the lengths of the observation and prediction trajectories, respectively. $(x_i^t, {y}_i^t)$ is the ground truth spatial coordinate of the $i^{th}$ pedestrian at time $t$. $(\hat{x}_i^t, \hat{y}_i^t)$ is the predicted future coordinate of the $i^{th}$ pedestrian at time $t$.

\begin{table*}[!t]
\centering
\caption{Comparison of LG-Traj (Our) with other approaches on ETH, HOTEL, UNIV, ZARA1, and ZARA2 datasets in terms of ADE/FDE (lower values are better). All approaches use the observed 8-time steps and predict the future 12-time steps. The top performance is highlighted in \textbf{bold}, and the second-best performance is indicated with \underline{underline}.}
\label{tab_eth_main}
\resizebox{\textwidth}{!}{%
\begin{tabular}{c|ccccccccc}
\hline 
Model & NMMP      & Social-STGCNN & PecNet          & Trajectron++ & SGCN       & STGAT     & CARPE           & AgentFormer           & GroupNet           \\ \hline
Venue & CVPR 2020 & CVPR 2020     & ECCV 2020       & ECCV 2020    & CVPR 2021  & AAAI 2021 & AAAI 2021       &  ICCV 2021         & CVPR 2022          \\ \hline
ETH   & 0.62/1.08 & 0.64/1.11     & 0.54/0.87       & 0.61/1.03    & 0.52/1.03  & 0.56/1.10 & 0.80/1.40       & 0.45/0.75          & 0.46/0.73          \\ \hline
HOTEL & 0.33/0.63 & 0.49/0.85     & 0.18/0.24       & 0.20/0.28    & 0.32/0.55  & 0.27/0.50 & 0.52/1.00       & 0.14/0.22          & 0.15/0.25          \\ \hline
UNIV  & 0.52/1.11 & 0.44/0.79     & 0.35/0.60       & 0.30/0.55    & 0.37/0.70  & 0.32/0.66 & 0.61/1.23       & 0.25/0.45          & 0.26/0.49          \\ \hline
ZARA1 & 0.32/0.66 & 0.34/0.53     & 0.22/0.39       & 0.24/0.41    & 0.29/0.53  & 0.21/0.42 & 0.42/0.84       &  0.18/0.30          & 0.21/0.39          \\ \hline
ZARA2 & 0.29/0.61 & 0.30/0.48     & 0.17/0.30       & 0.18/0.32    & 0.25/0.45  & 0.20/0.40 & 0.34/0.74       & 0.14/0.24          & 0.17/0.33          \\ \hline
AVG   & 0.41/0.82 & 0.44/0.75     & 0.29/0.48       & 0.31/0.52    & 0.37/0.65  & 0.31/0.62 & 0.46/0.89       & 0.23/0.39          & 0.25/0.44         \\ \hline \bottomrule
Model & GP-Graph  & STT           & Social-Implicit & BCDiff       & Graph-TERN & FlowChain & EigenTrajectory & SMEMO & Our                \\ \hline
Venue & ECCV 2022 & CVPR 2022     & ECCV 2022       & NIPS 2023    & AAAI 2023  & ICCV 2023 & ICCV 2023       & TPAMI 2024          & -                  \\ \hline
ETH   & 0.43/0.63 & 0.54/1.10   & 0.66/1.44       & 0.53/0.91    & 0.42/0.58  & 0.55/0.99 & 0.36/0.56      & 0.39/0.59          & 0.38/0.56 \\ \hline
HOTEL & 0.18/0.30 & 0.24/0.46   & 0.20/0.36       & 0.17/0.27    & 0.14/0.23  & 0.20/0.35 & 0.14/0.22      & 0.14/0.20          & 0.11/0.17 \\ \hline
UNIV  & 0.24/0.42 & 0.57/1.15   & 0.31/0.60       & 0.24/0.40    & 0.26/0.45  & 0.29/0.54 & 0.24/0.43       & 0.23/0.41          & 0.23/0.42 \\ \hline
ZARA1 & 0.17/0.31 & 0.45/0.94   & 0.25/0.50       & 0.21/0.37    & 0.21/0.37  & 0.22/0.40 & 0.21/0.39       & 0.19/0.32          & 0.18/0.33 \\ \hline
ZARA2 & 0.15/0.29 & 0.36/0.77   & 0.22/0.43       & 0.16/0.26    & 0.17/0.29  & 0.20/0.34 & 0.16/0.29       & 0.15/0.26          & 0.14/0.25 \\ \hline
AVG   & 0.23/0.39 & 0.43/0.88   & 0.33/0.67       & 0.26/0.44    & 0.24/0.38  & 0.29/0.52 & 0.23/0.38       & \underline{0.22/0.35}          & \textbf{0.20/0.34} \\ \hline 
\end{tabular}%
}
\end{table*}

\begin{table*}[!t]
\centering
\caption{Comparison of LG-Traj (Our) with other approaches on SDD dataset in terms of ADE/FDE (lower values are better). All approaches use the observed 8-time steps and predict the future 12-time steps. The top performance is highlighted in \textbf{bold}, and the second-best performance is indicated with \underline{underline}.}
\label{tab:sdd_main}
\resizebox{\textwidth}{!}{%
\begin{tabular}{c|c|c|c|c|c|c|c|c|c}
\hline 
Model & Trajectron++ & PECNet & SimAug & MG-GAN    & SGCN       & LBEBM  & PCCSNet & DMRGCN    & GroupNet        \\ \hline
Venue & ECCV 2020 & ECCV 2020 & ECCV 2020      & ICCV 2021      & CVPR 2021 & CVPR 2021        & ICCV 2021 & AAAI 2021  & CVPR 2022 \\ \hline
ADE   & 11.40        & 9.96   & 10.27  & 14.31     & 11.67      & 9.03   & 8.62    & 14.31   & 9.31      \\ \hline
FDE   & 20.12        & 15.88  & 19.71  & 24.78     & 19.10      & 15.97  & 16.16   & 24.78   &  16.11      \\ \hline \bottomrule
Model & CAGN         & SIT    & MID    & SocialVAE & Graph-TERN & BCDiff & MRL     & SMEMO       & Our            \\ \hline
Venue & AAAI 2022 & AAAI 2022 & CVPR 2022 & ECCV 2022 & AAAI 2023  & NeuIPS 2023 & AAAI 2023 & TPAMI 2024 & -         \\ \hline
ADE   & 9.42         & 9.13   & 9.73   & 8.88      & 8.43       & 9.05   & 8.22    & {\ul 8.11}  & \textbf{7.80}  \\ \hline
FDE   & 15.93        & 15.42  & 15.32  & 14.81     & 14.26      & 14.86  & 13.39   & {\ul 13.06} & \textbf{12.79} \\ \hline
\end{tabular}%
}
\end{table*}

\subsubsection{Implementation Details}
In our experiments, we chose different \textit{k} values for rank-\textit{k} trajectory augmentation. For ETH and HOTEL, we choose \textit{k=1}, while for UNIV, ZARA1, ZARA2, and SDD, \textit{k=3} is chosen. The token size is 16 for ETH and HOTEL, and 32 for UNIV, ZARA1, ZARA2, and SDD. We execute the experiments using Python 3.8.13 and PyTorch version 1.13.1+cu117. The training was conducted on NVIDIA RTX A5000 GPU with AMD EPYC 7543 CPU. Further details are provided in the supplementary material.

\begin{figure}
    \centering
    \includegraphics[scale=.062]{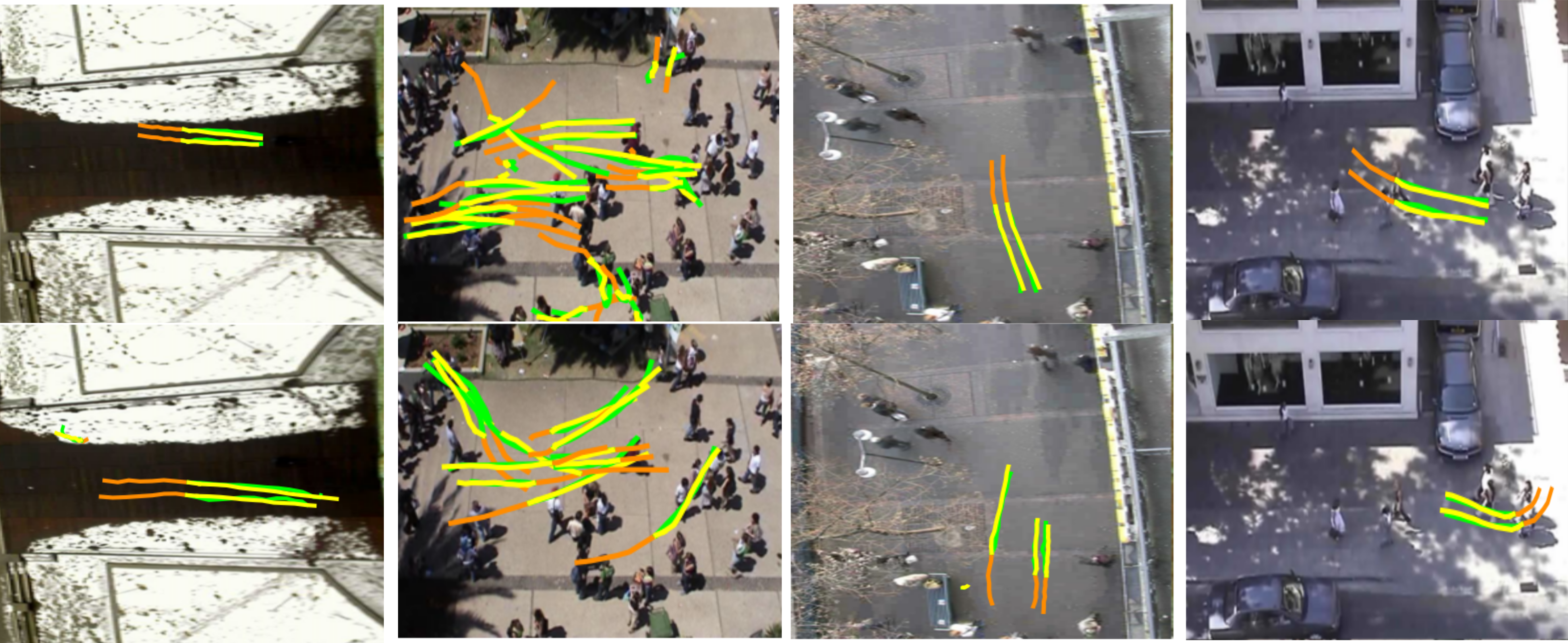}
   
\caption{Illustration of predicted trajectories from ETH (first column), UNIV (second column), HOTEL (third column), and ZARA (fourth column) datasets. Predicted pedestrian trajectories are highlighted in yellow. The observed trajectories are indicated in orange, while the ground truth trajectories are depicted in green. Our method demonstrates the prediction of future trajectories (yellow), closely matching the ground truth trajectories.}
\vspace{-4mm}
    \label{fig_traj_plot}
\end{figure}

\begin{table*}[]
\centering
\caption{Effect of Motion Cues (MC), Position Encoding (PE), and Trajectories Augmentation (TA) on model performance. Results show that all three components are crucial for accurate trajectory prediction.}
\label{tab:abl_compnent}
\resizebox{\textwidth}{!}{%
\begin{tabular}{c|c|c|c|c|c|c||c}
\hline 
Varients & ETH & HOTEL & UNIV & ZARA1 & ZARA2 & AVG & SDD \\ \hline
\textit{LG-Traj}                          & 0.38/0.56    & 0.11/0.17   & 0.23/0.42 & 0.18/0.33 & 0.14/0.25 & \textbf{0.20/0.34} & \textbf{7.80/12.79} \\ \hline
\textit{LG-Traj w/o MC}             & 0.49/0.87 & 0.45/0.92   & 0.35/0.67 & 0.28/0.56 & 0.22/0.43 & 0.36/0.69 & 22.1/41.2    \\ \hline
\textit{LG-Traj w/o PE}        & 0.42/0.62     & 0.12/0.17 & 0.24/0.44 & 0.19/0.35 & 0.14/0.26 & 0.22/0.36       & 8.23/13.41          \\ \hline
\textit{LG-Traj w/o TA} & 0.42/0.61    & 0.12/0.18   & 0.25/0.44 & 0.19/0.35 & 0.15/0.27 & 0.23/0.37         & 8.26/13.59          \\ \hline
\end{tabular}}

\end{table*}

\begin{table}[]
\centering
\caption{Comparing the performance with different token sizes for the tokenizer, the optimal token size is 16 for ETH and HOTEL datasets, while for UNIV, ZARA1, ZARA2, and SDD datasets, the best result is achieved with a token size of 32.}
\label{tab:abl_emb_size}
\begin{tabular}{c|c|c|c|c|c| |c}
\hline
Tokens & ETH & HOTEL & UNIV & ZARA1 & ZARA2  & SDD \\ \hline
\textit{16}                          & \textbf{0.38/0.56}    & \textbf{0.11/0.17}   & 0.24/0.43 & 0.18/0.33 & 0.14/0.25  & 7.85/12.90 \\ \hline
\textit{32}             & 0.41/0.60 & 0.12/0.18   & \textbf{0.23/0.42} & \textbf{0.18/0.33}  &  \textbf{0.14/0.25 }  & \textbf{7.78/12.79 }   \\ \hline
\textit{64}        & 0.42/0.62     & 0.12/0.18 & 0.24/0.45 & 0.19/0.45 & 0.14/0.26        & 7.80/12.86          \\ \hline
\end{tabular}%
\end{table}

\begin{table}[]
\centering
\caption{Information loss occurred (in \%) for different k values in rank-k approximation for trajectory augmentation.}
\label{tab:abl_informatin}

\begin{tabular}{c|c|c|c|c|c|c||c}
\hline
k-value & ETH & HOTEL & UNIV & ZARA 1 & ZARA 2 & AVG         & SDD         \\ \hline
\textit{k=1}     & 13\%  & 15\%     & 17\%    & 15\%      & 17\%      &15\%  & 27\%  \\ \hline
\textit{k=2}     & 7\%    & 9\%      & 10\%    & 8\%       & 8\%       & 8\%      &  15\%    \\ \hline
\textit{k=3}     & 3\%    & 2\%      & 4\%     & 3\%       & 3\%       & 3\%            & 6\%            \\ \hline
\end{tabular}%

\end{table}

\begin{figure*}
    \centering
    \includegraphics[scale=.085
]{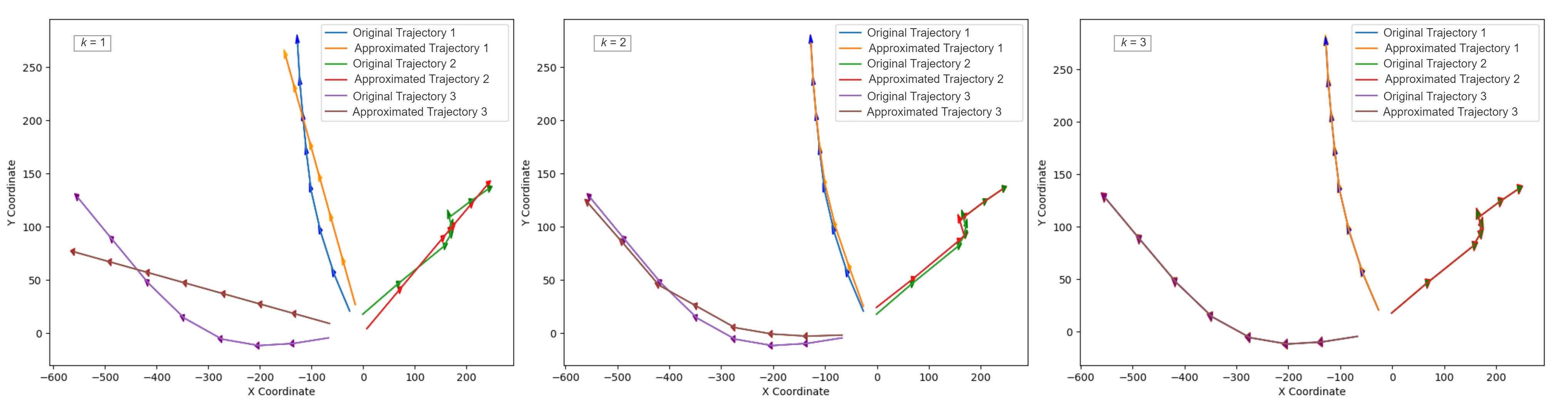}  
    \caption{Visualization of augmented trajectories for three pedestrians sampled from SDD using different \( k \) values in rank-\( k \) approximation.}
    \label{fig_turnacted}
\end{figure*}

\begin{figure}[!t]
    \centering
    \includegraphics[width=\columnwidth]{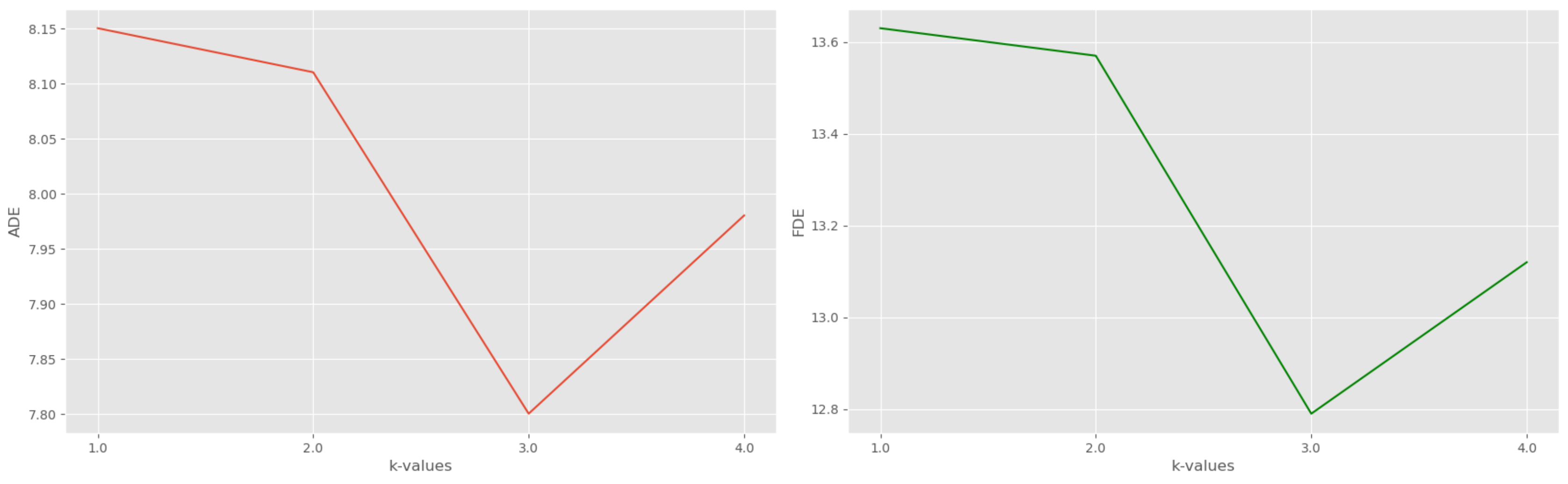}
    \caption{Visualization of ADE/FDE values obtained using our approach for different values of \textit{k} over the SDD dataset. The best result is obtained for \textit{k}=3.}
    \label{fig_k_plot}
\end{figure}

\subsection{Comparison with State-of-art Methods}

\subsubsection{Quantitative Results}

We compare our approach with recent methods. In Tables \ref{tab_eth_main}, we compare with SMEMO \cite{marchetti2024smemo}, FlowChain \cite{maeda2023fast}, Graph-TERN \cite{bae2023set}, EigenTrajectory \cite{bae2023eigentrajectory}, BCDiff \cite{li2024bcdiff}, Social-Implicit \cite{mohamed2022social}, STT \cite{monti2022stt}, GP-Graph \cite{bae2022gpgraph}, GroupNet \cite{Xu_2022_CVPR}, AgentFormer \cite{yuan2021agentformer}, CARPE \cite{mendieta2021carpe}, STGAT \cite{huang2019stgat}, SGCN \cite{shi2021sgcn}, Trajectron++ \cite{salzmann2020trajectron++}, PECNeT \cite{mangalam2020not}, Social-STGCNN \cite{mohamed2020social}, and NMMP \cite{hu2020collaborative}. Our model outperforms all compared methods in terms of average ADE/FDE on ETH-UCY and achieves a 10\%/3\%  relative improvement in average ADE/FDE compared to the recent method SMEMO \cite{marchetti2024smemo}. Similarly, in Table \ref{tab:sdd_main}, we present the experimental results on the SDD dataset in comparison with methods such as SMEMO \cite{marchetti2024smemo}, MRL \cite{wu2023multi}, BCDiff \cite{li2024bcdiff}, SocialVAE \cite{xusocialvae}, MID \cite{gu2022mid}, CAGN \cite{duan2022complementary}, STT \cite{monti2022stt}, GP-Graph \cite{bae2022gpgraph}, GroupNet \cite{Xu_2022_CVPR}, DMRGCN \cite{bae2021disentangled}, PCCSNet \cite{wu2023multi}, LBEBM \cite{pang2021lbebm}, SGCN \cite{shi2021sgcn}, MG-GAN \cite{dendorfer2021mg}, SimAug \cite{liang2020simaug}, PECNeT \cite{mangalam2020not}, and Trajectron++ \cite{salzmann2020trajectron++}. Our method outperforms all compared methods with ADE/FDE values of 7.80/12.79, respectively.

\subsubsection{Qualitative Results}

As shown in Figure~\ref{fig_traj_plot}, our approach predicts future trajectories that closely align with ground truth trajectories. The model trained using our approach effectively captures pedestrian interactions, movements, and various motion patterns present in the scene.

\subsection{Ablation Studies}

\subsubsection{Effect of various components in LG-Traj on model performance}
We investigate the impact of motion cues, position encoding, and trajectory augmentation on model performance. The results are presented in Table \ref{tab:abl_compnent}, with ADE/FDE values obtained by removing individual components from our approach. It is clear from the results that all three components are essential for our approach, with the most significant improvement achieved when motion cues are utilized. This demonstrates the significance of motion clues for the trajectory prediction task.

\subsubsection{Different token sizes for the tokenizer.}

In this ablation study, we investigate the impact of token size on trajectory prediction performance through experiments with varied token sizes, as reported in Table \ref{tab:abl_emb_size}. Our findings demonstrate that the optimal token size is 16 for the ETH and HOTEL datasets, while for the UNIV, ZARA1, ZARA2, and SDD datasets, the best results are achieved with a token size of 32.

\subsubsection{Information loss Vs. \textit{k} value}
We investigate the information loss (Eq. \ref{eq_info_loss}) incurred when augmenting the past trajectory using the $k$-rank approximation, and these losses are given in Table \ref{tab:abl_informatin}. Among the four singular values, the highest level of information loss is observed when \textit{k = 1}, while for \textit{k = 3} there is a minor information loss.

\subsubsection{Effect of using different \textit{k} values on performance} \label{sec_abl_inforloss}

Figure \ref{fig_turnacted} shows augmented trajectories for three pedestrians sampled from SDD for different $k$ values. Lower values of $k$ result in a significant loss of information, where only the direction of the trajectory is preserved. As the value of \textit{k} increases, more information is preserved in the augmented trajectories. For $k=3$, original and augmented trajectories exhibit a similar motion pattern on SDD. Figure \ref{fig_k_plot} shows ADE/FDE values obtained using our approach for different values of $k$ over the SDD dataset, and the best result is achieved when $k=3$.

\section{Conclusion}
In this work, we propose LG-Traj, a novel approach that capitalizes on past motion cues derived from a large language model (LLM) utilizing pedestrian past/observed trajectories. Furthermore, our approach integrates future motion cues extracted from pedestrian future trajectories through clustering of training data's future trajectories using a mixture of Gaussians. Subsequently, the motion encoder utilizes both past motion cues and observed trajectories, along with future motion cues, to model motion patterns. Finally, the social decoder incorporates social interactions among neighbouring pedestrians, along with the embedding produced by the motion encoder, to generate socially plausible future trajectories. Our experimental findings illustrate the feasibility of integrating LLM into trajectory prediction tasks. We showcase the effectiveness of our approach on widely used pedestrian trajectory prediction benchmarks, including ETH-UCY and SDD, and present various ablation experiments to validate our approach.

\clearpage  
\bibliographystyle{splncs04}
\bibliography{main}
\end{document}